\documentclass[conference]{IEEEtran}

\usepackage{amsmath,amssymb,amsfonts}
\usepackage{algorithmic}
\usepackage{graphicx}
\usepackage{textcomp}
\usepackage{multirow}
\usepackage{multicol}
\usepackage{xcolor}
\usepackage{xspace}
\usepackage{url}
\usepackage[bookmarks=true]{hyperref}
\usepackage{siunitx}
\usepackage[numbers, sort]{natbib}
\usepackage{threeparttable}
\usepackage{tikz}
\usepackage{flushend}
\usetikzlibrary{matrix}

\newcommand{\robot}{Berkeley Humanoid Lite\xspace}

\begin{document}

\title{Demonstrating \robot: \\ An Open-source, Accessible, and Customizable 3D-printed Humanoid Robot}

\author{Yufeng Chi, Qiayuan Liao, Junfeng Long, Xiaoyu Huang, \\
Sophia Shao, Borivoje Nikolic, Zhongyu Li, Koushil Sreenath \\ 
University of California, Berkeley
}

\maketitle

\begin{abstract}
Despite significant interest and advancements in humanoid robotics, most existing commercially available hardware remains high-cost, closed-source, and non-transparent within the robotics community. This lack of accessibility and customization hinders the growth of the field and the broader development of humanoid technologies. To address these challenges and promote democratization in humanoid robotics, we demonstrate Berkeley Humanoid Lite, an open-source humanoid robot designed to be accessible, customizable, and beneficial for the entire community. 
The core of this design is a modular 3D-printed gearbox for the actuators and robot body. All components can be sourced from widely available e-commerce platforms and fabricated using standard desktop 3D printers, keeping the total hardware cost under \$5,000 (based on U.S. market prices). 
The design emphasizes modularity and ease of fabrication.
To address the inherent limitations of 3D-printed gearboxes, such as reduced strength and durability compared to metal alternatives, we adopted a cycloidal gear design, which provides an optimal form factor in this context. 
Extensive testing was conducted on the 3D-printed actuators to validate their durability and alleviate concerns about the reliability of plastic components.
To demonstrate the capabilities of \robot, we conducted a series of experiments, including the development of a locomotion controller using reinforcement learning. These experiments successfully showcased zero-shot policy transfer from simulation to hardware, highlighting the platform's suitability for research validation.
By fully open-sourcing the hardware design, embedded code, and training and deployment frameworks, we aim for \robot to serve as a pivotal step toward democratizing the development of humanoid robotics. All resources are available at \url{https://lite.berkeley-humanoid.org}.
\end{abstract}

\IEEEpeerreviewmaketitle

\section{Introduction}
\label{sec:Introduction}

\begin{figure}[t]
\centering
\includegraphics[width=0.8\linewidth, trim={0 100 0 200}, clip]{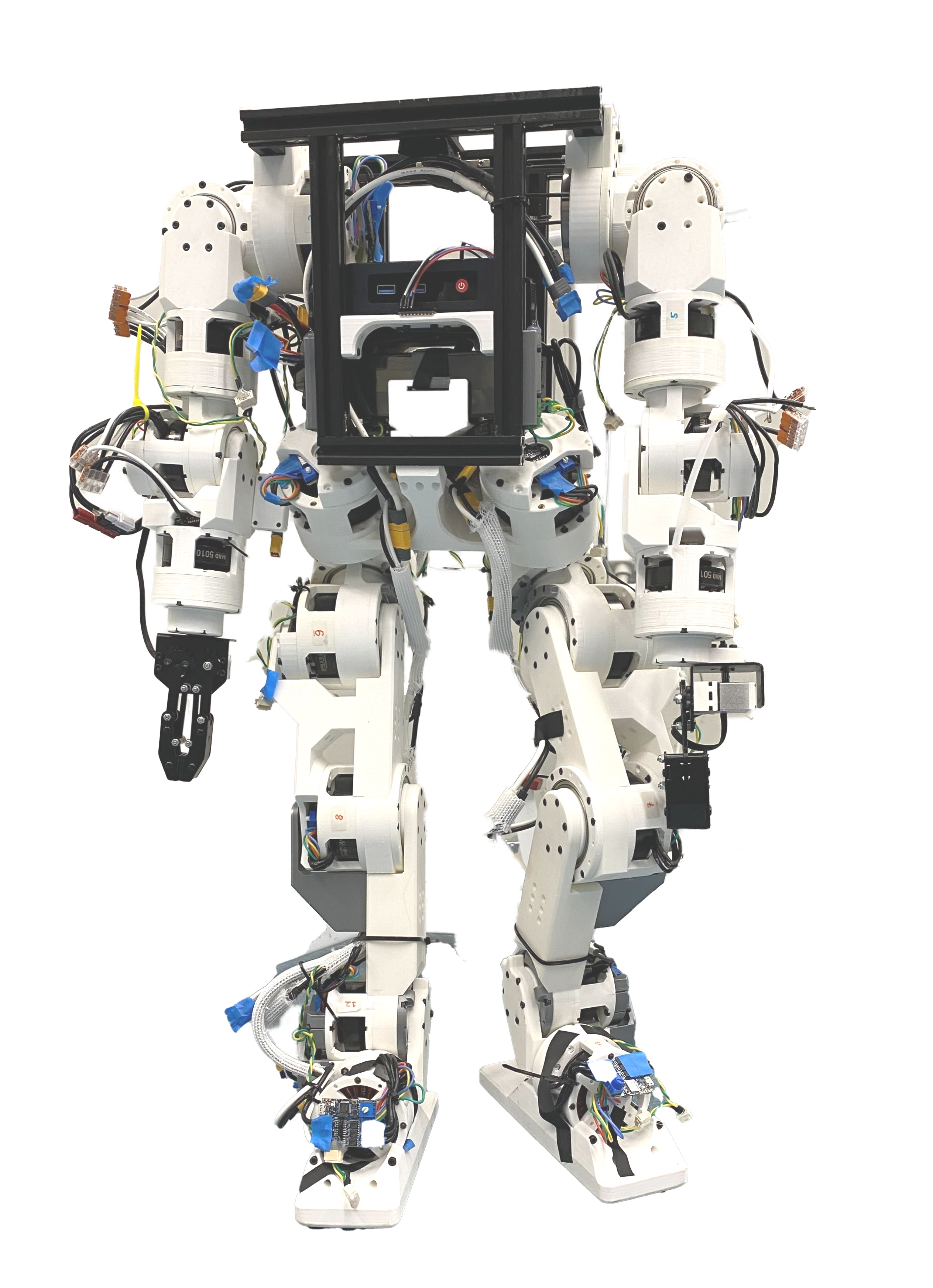}
\caption{Berkeley Humanoid Lite. An open-source, accessible, and customizable bipedal humanoid robot platform.}
\label{fig:berkeley-humanoid-lite}
\end{figure}

Humanoid robotics has gained substantial attention in research and industry, yet its accessibility remains a significant challenge. Most existing humanoid platforms are either commercial products with closed-source designs, high costs, and limited customizability, or research prototypes that require specialized manufacturing processes, making them inaccessible to the broader robotics community. This restricts innovation, as researchers, educators, and hobbyists face barriers in developing, modifying, and testing new ideas. Without affordable and customizable platforms, progress in fields such as reinforcement learning for bipedal locomotion, human-robot interaction, and robotic manipulation is hindered. Therefore, addressing the lack of open-source, cost-effective, and easily reproducible humanoid robot designs is crucial for democratizing robotics research and accelerating advancements in humanoid technologies.

Developing an open-source, cost-effective, and fully functional humanoid robot presents several challenges. First, humanoid robots require complex actuation mechanisms to achieve smooth and stable locomotion. While high-performance actuators exist, they are often expensive and proprietary, limiting accessibility. Prior research has explored various actuator designs, including planetary gear systems, belt-driven actuators, and quasi-direct drives, each with trade-offs in cost, efficiency, and complexity. However, few solutions provide both affordability and reliability while maintaining open-source availability.
Second, fabrication and assembly constraints further complicate accessibility. Many research-oriented humanoid platforms rely on CNC machining, laser cutting, or custom-milled components, which demand specialized manufacturing facilities. Open-source alternatives attempt to lower this barrier by utilizing 3D printing.
Finally, it remains difficult to ensure durable real-world performance. Humanoid robots must withstand continuous operational stresses while maintaining precise motion control. The use of 3D-printed components, while improving accessibility, introduces durability concerns due to material limitations such as reduced stiffness and wear over time. 
Thus, designing a humanoid platform that balances accessibility, modularity, and long-term reliability remains an open challenge in robotics research.

To address these challenges, we introduce Berkeley Humanoid Lite, an open-source, mid-scale humanoid robot platform designed for accessibility, affordability, and customizability. Our approach leverages 3D-printed cycloidal gear actuators, which balance cost-effectiveness with sufficient mechanical robustness, reducing dependency on expensive, proprietary actuators. By optimizing the actuator design and employing widely available components, our platform ensures that researchers and hobbyists can build and modify the robot without requiring specialized manufacturing facilities.
To enhance fabrication accessibility, we focus on modular and scalable design principles. All structural components can be 3D-printed using standard desktop printers, and all electronic and mechanical parts can be sourced from common online vendors. This approach significantly lowers the barriers to entry for constructing and maintaining a humanoid robot, making it feasible for a wider range of users.
Furthermore, we demonstrate the real-world viability of our platform through experiments in locomotion and teleoperation. By implementing a reinforcement learning-based locomotion controller, we achieve zero-shot policy transfer from simulation to hardware, validating the effectiveness of our actuator design in dynamic tasks. Additionally, the teleoperation system enables real-time manipulation tasks, further showcasing the robot’s adaptability for research and education.
By making the hardware design, embedded system code, and training frameworks fully open-source, we aim to democratize humanoid robotics research. Our contribution is to provide an accessible, customizable, and cost-effective platform, encouraging broader participation in humanoid development and innovation.

The main contributions of this work are summarized as follows: (1) We introduce a humanoid robot platform that is accessible and customizable. (2) We demonstrate the platform’s capability for locomotion and whole-body control tasks. (3) We provide a fully open-source hardware and software stack with accompanying detailed instructions, enabling researchers and enthusiasts worldwide to replicate, customize, and improve upon our platform.

\section{Related Works}
\label{sec:Related Works}

Our focus is on open-source, accessible, and customizable actuators for robotics applications, as well as robot platforms that can be easily adopted by academic labs and research groups.

\subsection{Actuators}

Actuators play a crucial role in robotic platforms, directly impacting performance, efficiency, and capabilities. Various design configurations have been explored to optimize robotic actuators.
\citet{katz2018low} developed a fully open-source, high-performance, proprioceptive actuator, including a custom motor driver for quadrupedal robots. Their design utilizes an off-the-shelf motor and a planetary gear component set, requiring modifications to both the motor and gears, along with a custom aluminum housing. \citet{gealy2019quasi} employed belt reduction systems as modular actuators for the arm. Similarly,  \citet{kau2019stanford} demonstrated promising results in quadrupedal robotics by utilizing belt drive and ODrive~\cite{odriverobotics} as the motor controller, however, their design features a low gear ratio and lacks modularity. In addition, \citet{grimminger2020open} and \citet{wuthrich2020trifinger} presented a series of open-source works incorporating belt reduction and custom motor drivers for small-scale legged robots.
\citet{urs2022design} introduced a 3D-printed planetary gearbox combined with the Moteus~\cite{moteus_r4_11} motor driver, though their approach requires a resin 3D printer. \citet{azocar2018design} developed fully open-source actuators with planetary gearing for prosthetic applications, but their method necessitates CNC machining.
Furthermore, \citet{roozing20223d} investigated the design considerations for manufacturing cycloidal gearboxes using Fused Deposition Modeling (FDM) 3D printing. In a subsequent study, \citet{roozing2024experimental} proposed a non-pinwheel design, optimizing the form for 3D-printed actuators. Their research highlights the feasibility of integrating 3D-printed cycloidal drives into actuators, paving the way for low-cost, accessible, and reliable robotic systems.

\subsection{Humanoids}

Existing humanoid robot platforms generally fall into three categories: commercial proprietary systems, research lab designs, and open-source community projects.

Commercial platforms—such as Agility Robotics’ Digit~\cite{agility_robotics_digit}, Fourier Intelligence’s GR1~\cite{fourier_gr1}, Westwood Robotics’ THEMIS~\cite{ahn2023development}, OpenLoong FXV2750~\cite{openloong_fxv2750}, and NAO H25~\cite{nao_h25}—are engineered for robust execution. However, their substantial price tag renders them largely inaccessible to most research groups and the broader robotics community. Although lower-cost alternatives like Unitree’s H1~\cite{unitree_h1} and G1~\cite{unitree_g1} are available, their closed-source architecture restricts user modifications and complicates repairs, limiting customization for research applications.

Research labs have also been developing in-house humanoid robots that offer high degrees of freedom (DoF) and substantial torque, serving as advanced platforms for investigating humanoid robot control~\cite{lohmeier2009humanoid, kaneko2019humanoid, park2005mechanical}. More recently, smaller-scale robots—such as  Berkeley Humanoid~\cite{liao2024berkeley}, MIT Humanoid~\cite{saloutos2023design}, Wukong~\cite{wang2022hybrid} and many others~\cite{zhu2023design, liu2022design, li2023dynamic, xia2024duke, shi2025toddlerbot}—have been introduced with a focus on agile locomotion. Although these platforms achieve high performance and are well-suited for reinforcement learning applications, their limited production quantities typically confine their use to a few institutions. Furthermore, these designs often depends on advanced manufacture techniques such as CNC machining and laser cutting, which are only available at select academic institutes. Expensive components are also used which restricts adaptability and broader adoption. As a result, the need for specialized manufacturing facilities and expertise significantly limits the accessibility of these platforms to the wider research community and hobbyists.

Open-source community projects offers a more accessible alternative. However, many still depend on custom parts that requires CNC manufacturing~\cite{parmiggiani2012design}. Other projects, such as the Poppy Humanoid~\cite{lapeyre2014rapid}, OP3~\cite{robotis_op3}, and NimbRo-OP2~\cite{ficht2017nimbro}, take advantage of 3D printing and utilize servo motors as the joints. This approach lowers the manufacturing barrier by enabling fabrication with desktop 3D printers. Nevertheless, servos typically lack backdrivability and exhibit high reflected inertia, making them slower and less suitable for scaling up to mid-scale or larger humanoid platforms.

To address these challenges, we introduce \robot, a modular, mid-scale humanoid robot platform that emphasizes both accessibility and customization. \robot is designed to lower the barrier to entry, enabling a broader community of researchers and hobbyists to engage in humanoid robotics research and innovation.

\section{System Design}

\subsection{Overview}
\begin{figure}[t]
  \centering
  \includegraphics[width=0.4\textwidth]{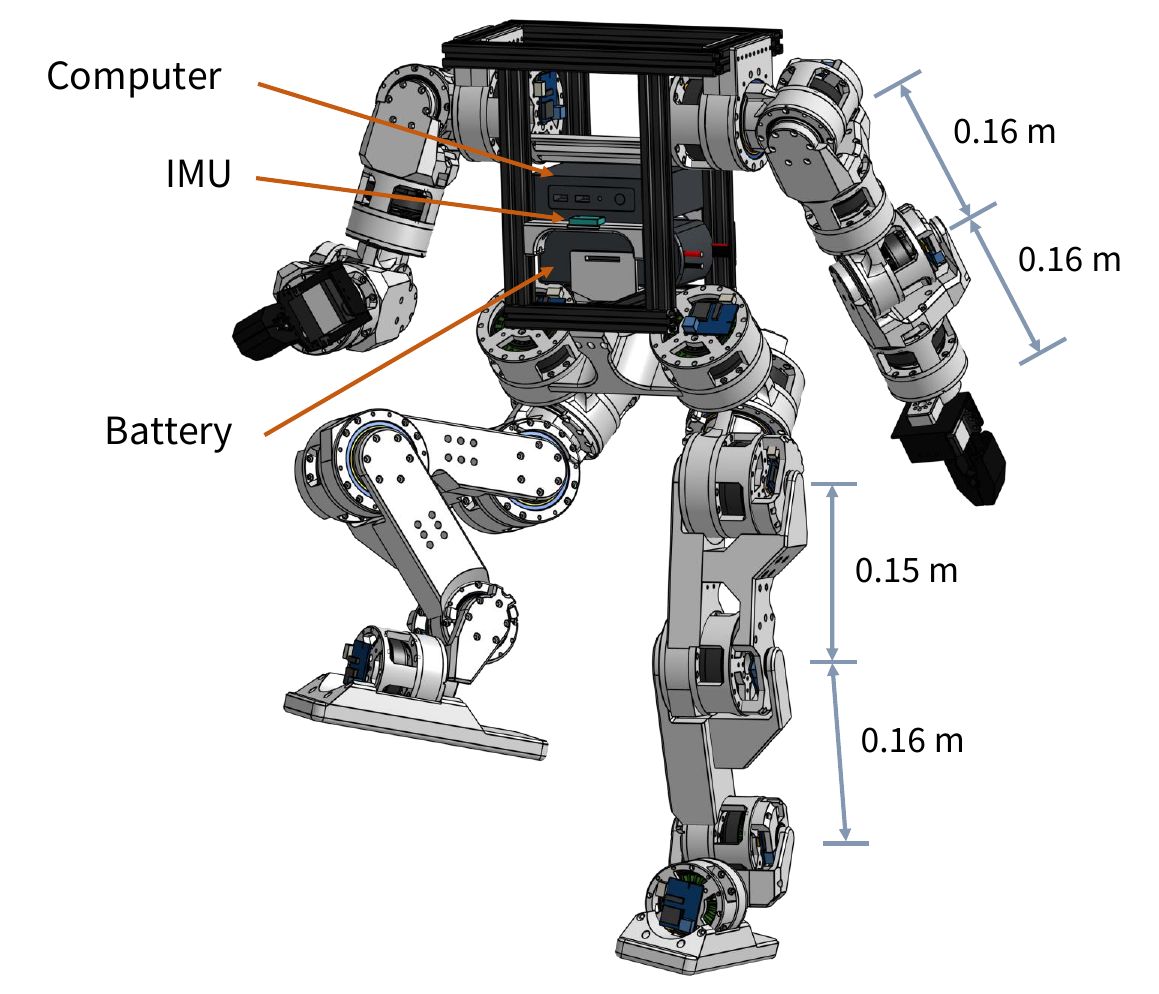}
  \caption{\label{fig:system-components}Main components and key dimensions of the limbs of the \robot.}
\end{figure}

The \robot is a mid-scale humanoid robot platform that is designed with accessibility and customizability in mind. The main components of the robot are shown in Figure~\ref{fig:system-components}. The robot weighs 16~\si{kg} and stands 0.8~\si{m} tall. Two sizes of actuators, each incorporating a 3D-printed cycloidal gearbox, directly drive the joints in the legs and arms. The torso is built with aluminum extrusions to support mounting additional components, and a cellphone-grade IMU is positioned near the center of the torso.

High-speed and low-latency communication between the actuators, the IMU, and the controller is crucial to achieve smooth motion. An Intel N95 mini PC located at the center of the torso serves as the low-level control computer. Each limb's joint actuators are connected via a 1~\si{Mbps} CAN 2.0 bus and the four buses interface to the computer through USB-CAN adapters. The IMU is connected over USB through an Arduino microcontroller. The communication rate to both the actuators and IMU is configured to be 250~\si{Hz}. RL-based locomotion policies are also deployed on the same computer, with details provided in Section \ref{sec:locomotion}.

The robot is powered by an onboard 6S 4000 mAh Lithium Polymer (LiPo) battery that provides approximately 30 minutes of operation. Tethering to an external power supply is also supported for extended testing sessions.

Apart from the non-standard structural parts, which are printed on a common desktop 3D printer, all components can be purchased from major online vendors in multiple countries. These design choices reflect the goal of creating an accessible and customizable humanoid platform.

\begin{figure}[t]
  \centering
  \includegraphics[width=\linewidth]{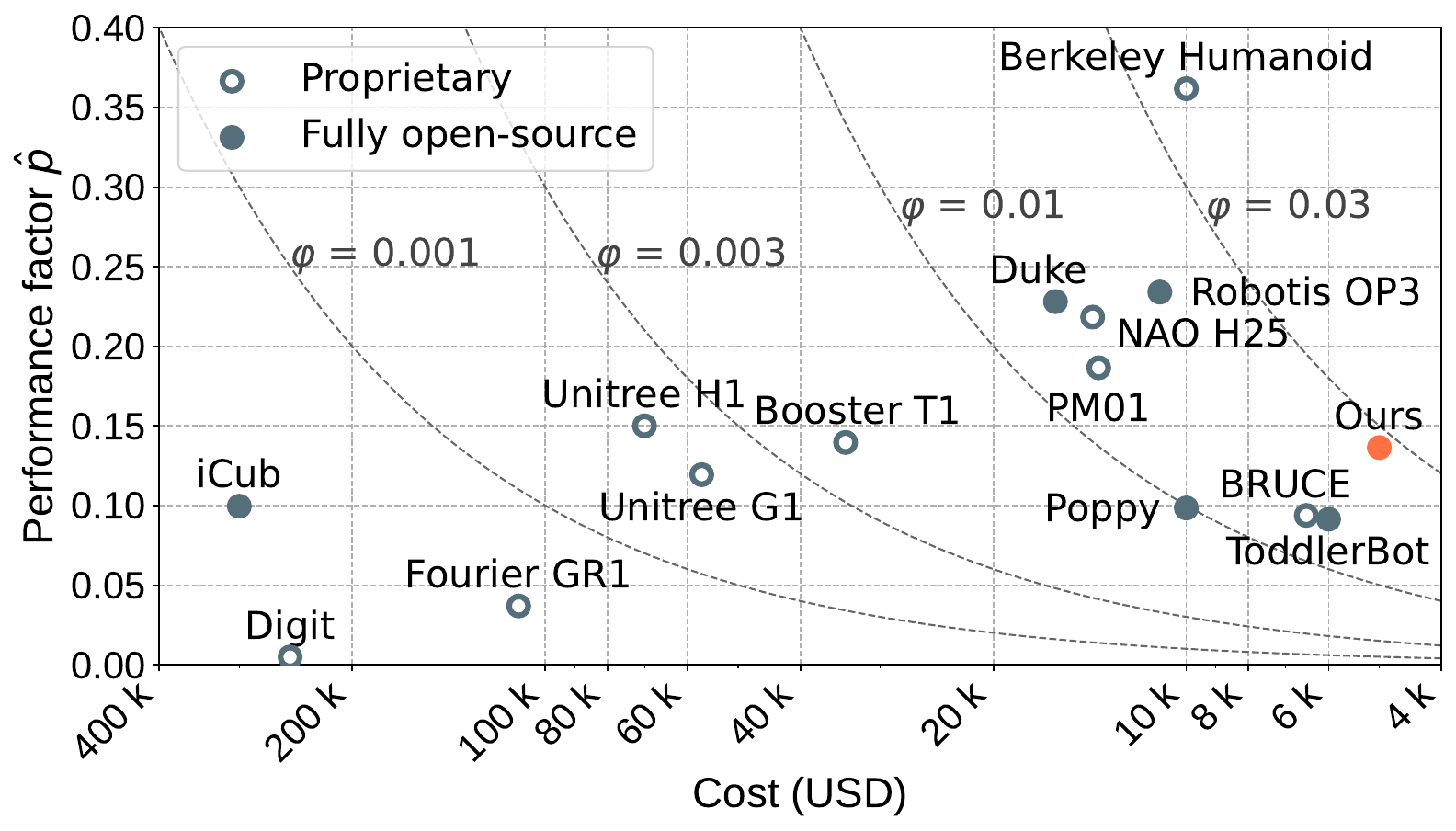}
  \caption{Comparison between existing robot platforms. Y-axis shows the performance factor $\hat{p}$ defined in \eqref{eqn:performance-factor}, representing the average peak torque normalized by the weight and height of the robot. A higher performance-per-dollar $\varphi$, defined in \eqref{eqn:performance-factor-per-dollar}, indicates better cost-effectiveness, and open-source hardware and software platform enables customization towards specific user requirements.}
  \label{fig:comparision}
\end{figure}

\subsection{Performance Factor}

To be able to benchmark against other robots and illustrate our focus on accessibility and customizability while maintaining sufficient performance, 
we adopt the quantitative performance metric proposed by~\citet{shi2025toddlerbot}, with a minor modification. Specifically, we normalize the metric by the number of joints, and define the performance factor as the average peak torque of all actuated DoFs, normalized by the robot's height and weight: 
\begin{equation}
\label{eqn:performance-factor}
\hat{p} = \frac{1}{N h mg} \sum_{i=1}^N |\tau_i^{\max}|,
\end{equation}
where $N$ denotes the number of actuated DoFs, $h$ and $mg$ represent the height and weight of the robot, and $|\tau_i^{max}|$ represents the maximum torque of the i-th joint motor. To incorporate cost-effectiveness into consideration, the performance-per-dollar is then defined as the performance factor divided by the cost or selling price of the robot:
\begin{equation}
\label{eqn:performance-factor-per-dollar}
\varphi = \frac{\hat{p}}{\text{cost}}.
\end{equation}
As shown in Fig.~\ref{fig:comparision}, our platform achieves a high performance factor with a cost lower than \$5000.

\begin{figure}[t]
    \centering
    \includegraphics[width=0.45\textwidth]{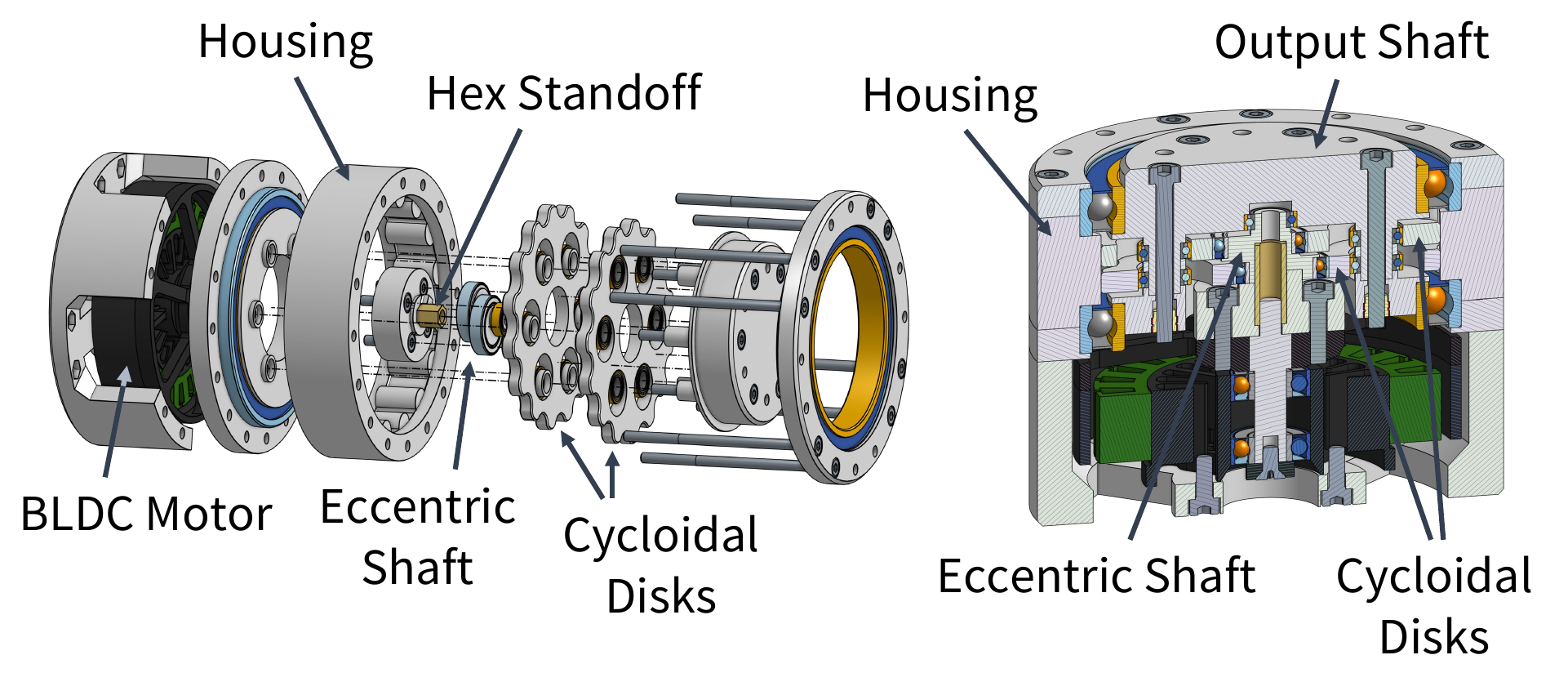}
    \caption{Exploded and cross-sectional views of the 6512 actuator. All of the components are either 3D-printed or sourced from online marketplaces.}
    \label{fig:6512-cross-section-explode-view}
\end{figure}

\subsection{Design Towards Accessibility}

A key priority in developing \robot is to make the hardware platform accessible to a wide range of users. To realize this goal, the design process emphasizes widely available components, straightforward fabrication methods, and overall cost control.

\subsubsection{Component Availability}

Many open-source robot designs depend on specialized components, such as custom thrust bearings or high-performance motor drivers, which can be difficult to source. In contrast, \robot relies on parts from commonly used online marketplaces, notably Amazon and Taobao (AliExpress), for most off-the-shelf elements. Electronic components are obtained from reputable suppliers such as Digikey and Mouser. By selecting parts that are broadly available in multiple countries, \robot minimizes supply barriers and ensures that users can acquire necessary materials without extensive lead times or geographic restrictions.

\subsubsection{3D Printing}

Recent advances in desktop 3D printing technology have made Fused Deposition Modeling (FDM) more accessible and reliable. Accordingly, \robot employs FDM to fabricate all non-standard structural parts. Polylactic Acid (PLA) is selected as the filament material for its robust mechanical properties and favorable printing characteristics. Although 3D-printed plastic parts introduce challenges—such as lower material strength, manufacturing variances, and anisotropic properties—\robot mitigates these concerns by adopting cycloidal gearboxes in its actuators. Cycloidal gears distribute loads over multiple teeth and accommodate the limited resolution of desktop 3D printers more effectively than planetary gears, thus enhancing reliability~\cite{roozing20223d}. As shown in Figure~\ref{fig:6512-cross-section-explode-view}, additional reinforcements, such as through-screws and embedded brass hex stands, help prevent failures along layer boundaries and bolster overall load capacity. Certain subcomponents are merged into single 3D prints to reduce weight and avoid potential stress concentrations around fasteners. To maintain compatibility with standard desktop machines, every part is designed to fit within a 200~\si{mm} × 200~\si{mm} × 200~\si{mm} build volume.

\subsubsection{Low Cost}

We detail the component costs in the United States and China to illustrate the affordability of \robot. Tables~\ref{table:6512-bom-cost} and \ref{table:5010-bom-cost} summarize the expenses for the two actuator designs, while Table~\ref{table:humanoid-bom-cost} presents the total cost for the assembled robot. Beyond raw materials, the time for building and repairing the robot also factors into overall expense. For \robot, most off-the-shelf parts ship within a week in both the United States and China, custom parts can be printed within a week, and the entire robot can be assembled in about three days. Broken actuators can also be easily swapped out and repaired. These considerations minimize repair downtime and lower the barrier to entry for constructing and maintaining a mid-scale humanoid robot.

As a concrete example, for the 6512 actuator, we chose the widely available 6811ZZ ball bearing to define the overall actuator size. The M6C12 150KV BLDC drone motor from MAD Components is used to drive the actuator, which strikes a good balance between availability, performance, and cost. We then opted for the B-G431B-ESC1 as the motor driver, favoring its affordability and ready availability at Mouser and Digikey. Although alternatives like the Moteus Controller~\cite{moteus_r4_11}, ODrive~\cite{odriverobotics}, or VESC~\cite{vesc_project} can also be used, we prioritized cost-effectiveness and consistent supply. The housing, cycloidal gear, input shaft, and output shaft are all 3D printed, with a brass hex stand embedded in the input shaft to boost stiffness and improve torque transfer from the motor to the cycloidal disks.

\begin{table}[t]
  \centering
  \caption{Bill-Of-Materials (BOM) of the 6512 Actuator}
  \label{table:6512-bom-cost}  
  \begin{threeparttable}
    \begin{tabular}{ |c|c|c| }
    \hline
    Item                      & Cost (US) & Cost (China) \\
    \hline
    M6C12 BLDC Drone Motor    & \$129  & \$124  \\
    \hline
    B-G431B-ESC1 Motor Driver &  \$19  & \$23   \\
    \hline
    AS5600 Position Encoder   &   \$3  & \$1    \\
    \hline
    Bearings                  &  \$23  & \$4    \\
    \hline
    Fasteners                 &   \$5  & \$1    \\
    \hline
    3D Printed Parts          &   \$4  & \$1    \\
    \hline
    Misc (Cables, Connectors) &   \$5  & \$3    \\
    \hline
    Total                     & \$188  & \$157 \\
    \hline
    \end{tabular}
    \begin{tablenotes}
      \item All values are rounded to the nearest dollar.
    \end{tablenotes}
  \end{threeparttable}
\end{table}

\begin{table}[t]
  \centering
  \caption{Bill-Of-Materials (BOM) of the 5010 Actuator}
  \label{table:5010-bom-cost}
  \begin{threeparttable}
    \begin{tabular}{ |c|c|c| }
    \hline
    Item                      & Cost (US) & Cost (China) \\
    \hline
    5010 BLDC Drone Motor     & \$84  & \$62  \\
    \hline
    B-G431B-ESC1 Motor Driver & \$19  & \$23  \\
    \hline
    AS5600 Position Encoder   & \$3   & \$1   \\
    \hline
    Bearings                  & \$18  & \$3   \\
    \hline
    Fasteners                 & \$4   & \$1   \\
    \hline
    3D Printed Parts          & \$3   & \$1   \\
    \hline
    Misc (Cables, Connectors) & \$5   & \$3   \\
    \hline
    Total                     & \$136 & \$94  \\
    \hline
    \end{tabular}
    \begin{tablenotes}
      \item All values are rounded to the nearest dollar.
    \end{tablenotes}
  \end{threeparttable}
\end{table}

\begin{table}[t]
  \centering
  \caption{Bill-Of-Materials (BOM) of a Humanoid Robot}
  \label{table:humanoid-bom-cost}
  \begin{threeparttable}
    \begin{tabular}{ |c|c|c| }
    \hline
    Item                         & Cost (US) & Cost (China) \\
    \hline
    Mini PC                      & \$129    & \$223  \\
    \hline
    USB-CAN Adapters (4x)        & \$68     & \$43   \\
    \hline
    USB Hubs (2x)                & \$36     & \$11   \\
    \hline
    BNO085 IMU                   & \$13     & \$12   \\
    \hline
    6S LiPo Battery              & \$70     & \$81   \\
    \hline
    6512 Actuators (10x)         & \$1,880  & \$1,563 \\
    \hline
    5010 Actuators (12x)         & \$1,632  & \$1,130 \\
    \hline
    Grippers (2x)                & \$72     & \$44   \\
    \hline
    Aluminum Extrusions          & \$39     & \$3    \\
    \hline
    3D Printed Components        & \$200    & \$84   \\
    \hline
    Misc. Structural Components  & \$50     & \$14   \\
    \hline
    Misc. Electronic Components  & \$123    & \$28   \\
    \hline
    Total                        & \$4,312  & \$3,236 \\
    \hline
    \end{tabular}
    \begin{tablenotes}
      \item All values are rounded to the nearest dollar.
    \end{tablenotes}
  \end{threeparttable}
\end{table}

\begin{figure}[t]
  \centering
  \includegraphics[width=0.45\textwidth]{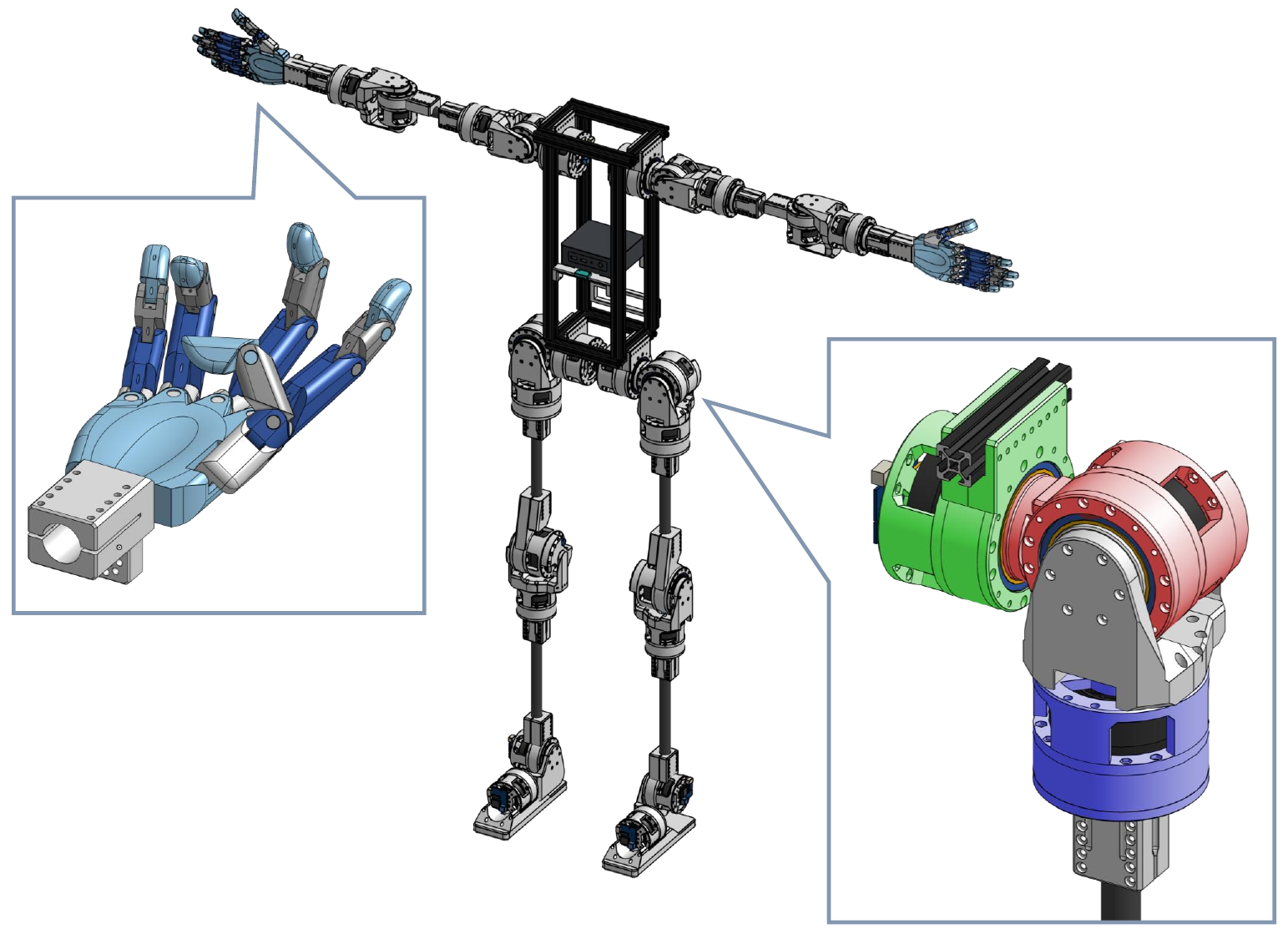}
  \caption{An alternative adult-sized robot configuration with 7 degrees-of-freedom leg and dexterous hand. Carbon fiber tubes that are also available from online vendors are used to increase the length of the legs to match an average adult. Colors are used to show joint orientation (green: thigh pitch; red: abduction; blue: thigh rotation).}
  \label{fig:different-configurations-full-size-robot}
\end{figure}

\subsection{Design Towards Customization}

In addition to accessibility, \robot is designed to be highly customizable. Different research objectives and tasks may require variations in robot dimensions and joint configurations, and the flexibility of 3D printing makes it straightforward to adapt the robot’s morphology to suit specific needs.

\subsubsection{Flexible Configurations}

Since each joint in \robot is driven by a self-contained actuator with no additional linkages, adjusting the link length between adjacent joints is straightforward. 
As illustrated in Figure~\ref{fig:different-configurations-full-size-robot}, extending the leg links with carbon fiber tubes transforms \robot into an adult-scale platform.
Joint orders are also easily interchangeable: while most humanoid robots use three consecutive revolute joints at the hip to approximate a ball joint, variations in joint arrangement and orientation can significantly influence range of motion, power consumption, and efficiency~\cite{zorjan2011influence, zorjan2015influence}. Figure~\ref{fig:different-configurations-full-size-robot} also illustrates an alternative hip configuration that is commonly employed in other humanoids. By streamlining these modifications, \robot enables rapid prototyping and comparative evaluation of different hip designs on real hardware.

\subsubsection{Different Morphologies}

\begin{figure}[t]
  \centering
  \includegraphics[width=0.45\textwidth]{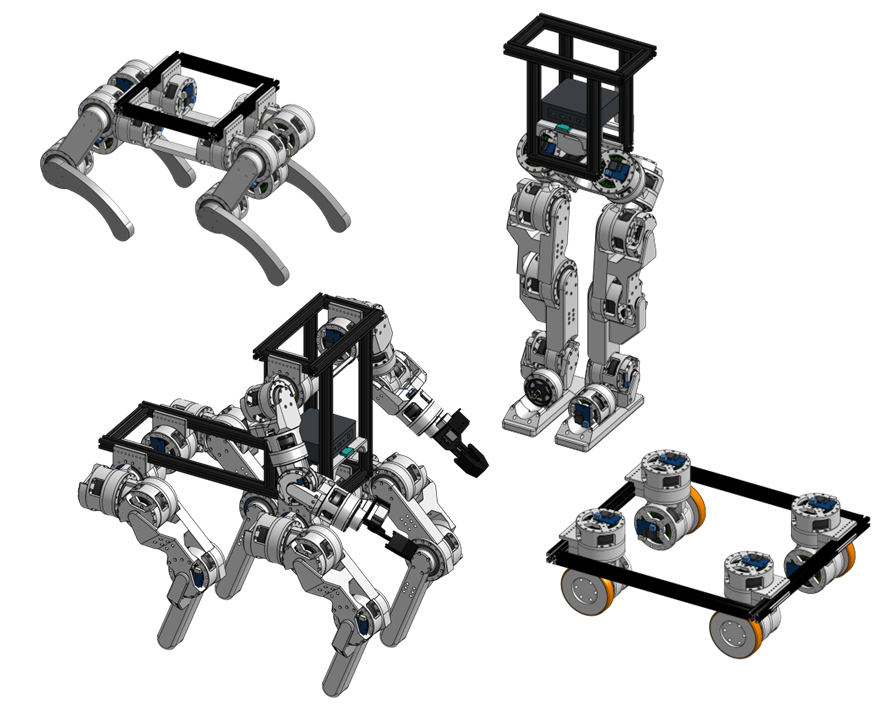}
  \caption{\label{fig:different-configurations-morphs} With minimal modifications, the modular joint actuators can be arranged to form different robot morphologies. Example forms include quadruped (top-left), biped (top right), Centaur-like (bottom-left), and mobile base (bottom-right).}
\end{figure}

Berkeley Humanoid Lite’s actuators require only power and CAN communication, which allows each CAN bus to support up to 64 devices. This benefit makes it easy to reconfigure the robot into a variety of morphologies. 
Figure~\ref{fig:different-configurations-morphs} highlights examples of a quadruped, a biped, a Centaur-like arrangement, and a mobile base platform, illustrating how the 3D-printed structure and self-contained actuators can be adapted for diverse robotic designs with minimal changes.

\section{Actuator Evaluation}

A reliable actuator is fundamental to the robot’s overall performance. To assess the capabilities of our proposed design, we conducted a set of experiments under conditions identical to those on the robot, including a 24~\si{V} power supply, identical position PD gains, and matching position, torque, and current bandwidth configurations. For further verification, we cross-validated the outcomes obtained with our custom motor controller and firmware against those produced by an off-the-shelf Moteus Controller.

\subsection{Power Efficiency}

\begin{figure}[t]
  \centering
  \includegraphics[width=0.45\textwidth]{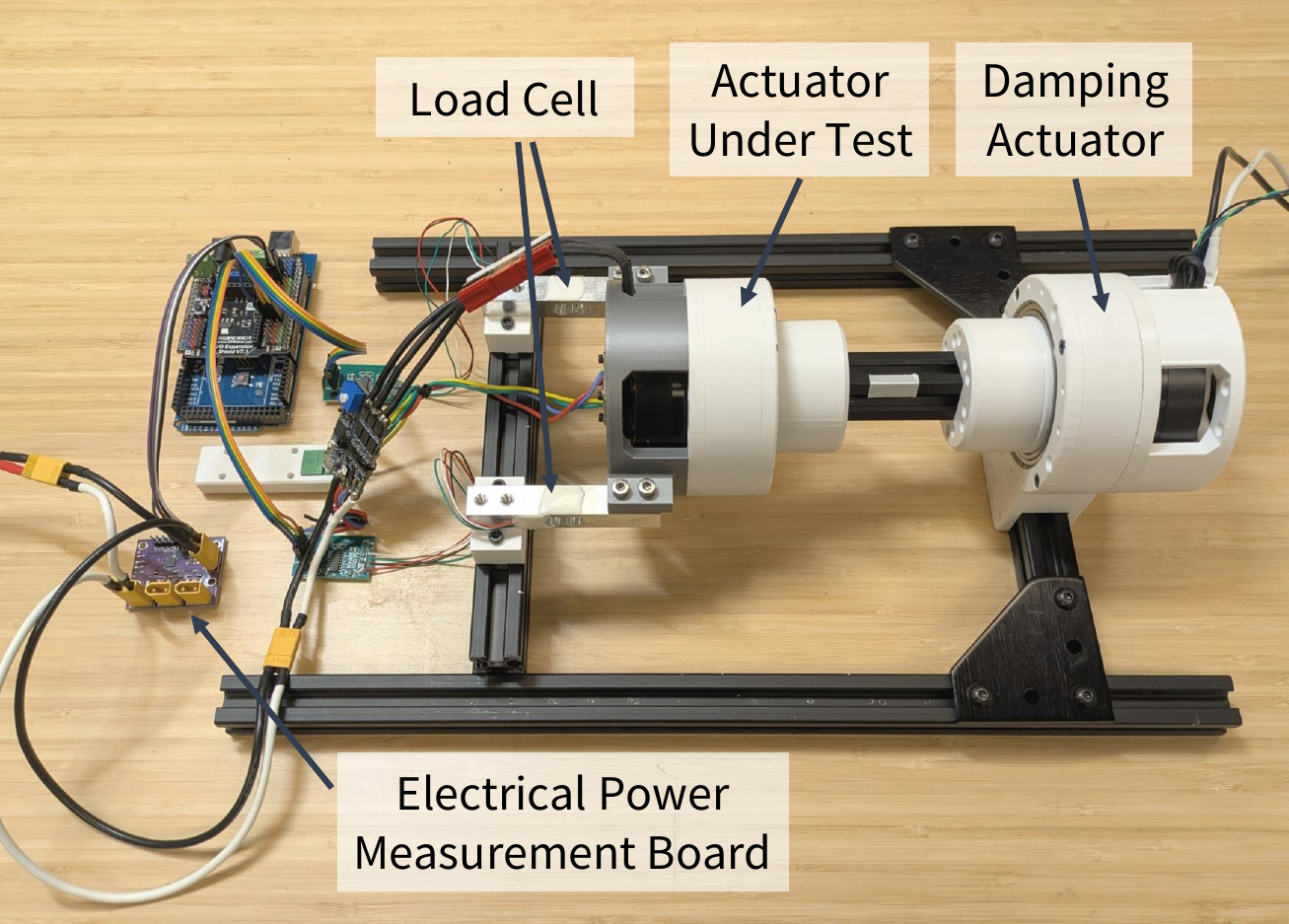}
  \caption{Power efficiency test setup. Two load cells are used to directly measure the torque of the actuator under test, while another actuator acts as a damper, absorbing the mechanical energy produced by the actuator under test.}
  \label{fig:experiment-test-setup}
\end{figure}

To evaluate the actuator’s power efficiency, we employed a custom dynamometer stand (Figure~\ref{fig:experiment-test-setup}). The actuator under test was operated in torque-control mode, while a secondary damping actuator—run in velocity-control mode—held the system at a constant rotational speed. Torque produced by the primary actuator was measured with two load cells. An electrical-power measurement board logged the supply voltage and current, from which electrical input power was calculated. Mechanical output power was obtained by multiplying the measured torque by the measured rotational velocity. Mechanical efficiency is defined as the ratio of the measured mechanical power to the absolute value of the product of the commanded torque and velocity, which represents the efficiency of the cycloidal gear reducer. Total efficiency was defined as the ratio of the measured mechanical power to the electrical input power, thereby reflecting the overall actuator efficiency, which includes motor copper losses, driver electrical losses, and mechanical losses.

\begin{figure}[t]
  \centering
  \includegraphics[width=0.9\linewidth]{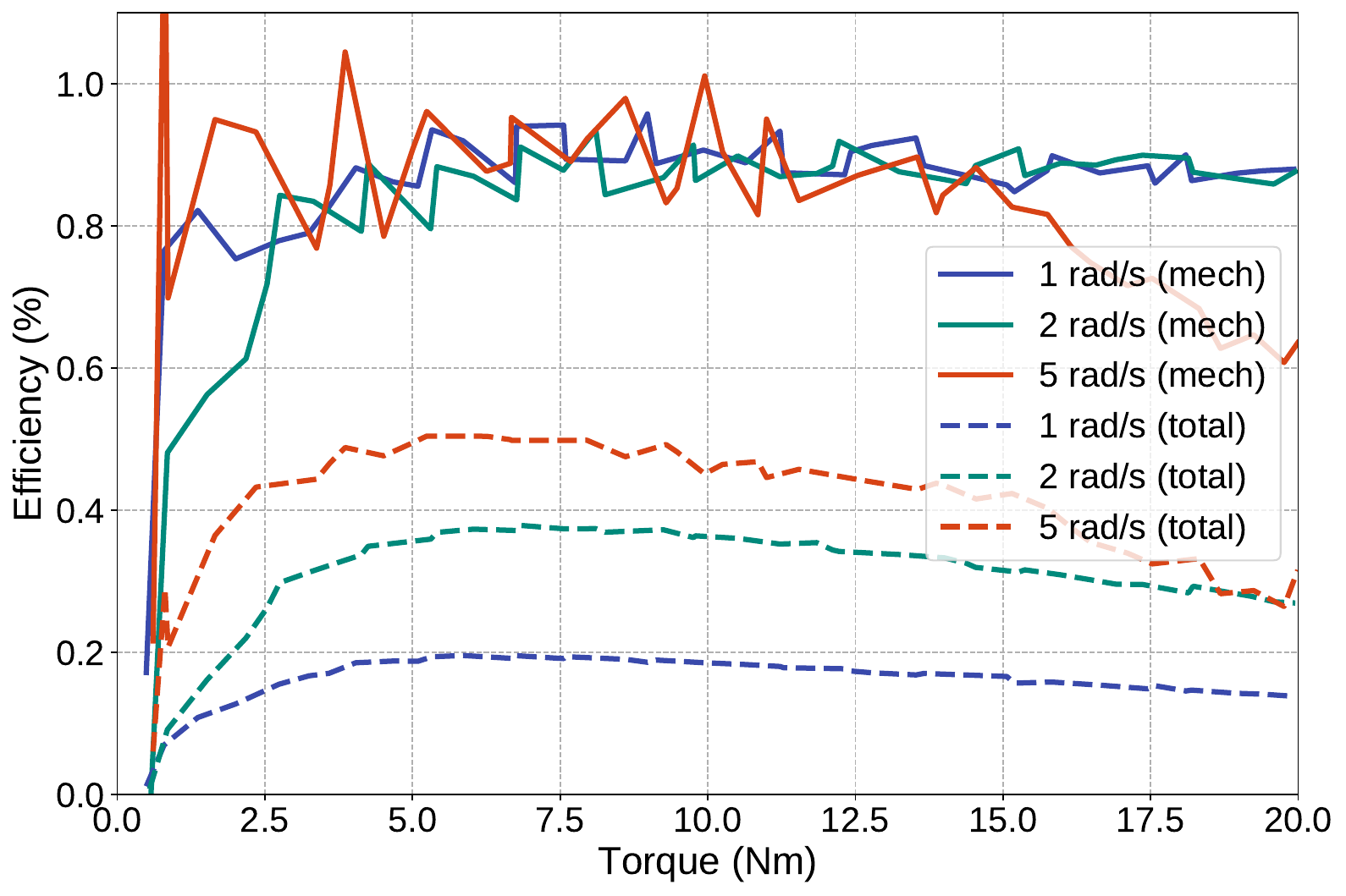}
  \centering
  \includegraphics[width=0.9\linewidth]{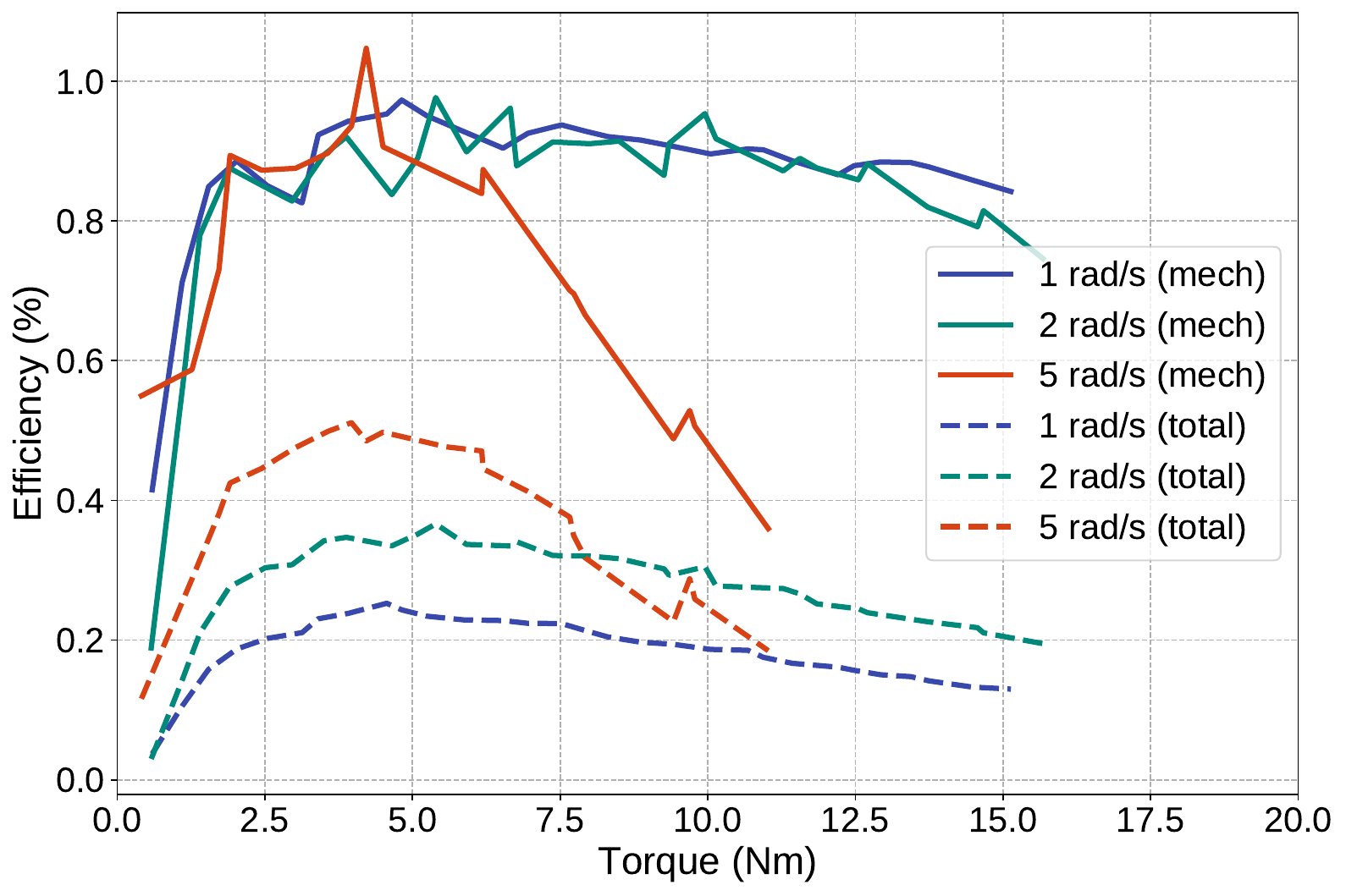}
  \caption{Efficiency of the 6512 actuator (top), and the 5010 actuator (bottom). Mechanical efficiency is calculated as the ratio between the measured mechanical power to the product of the commanded torque and velocity, whereas total efficiency is defined as the ratio of the measured mechanical power to the electrical input power. The results indicate that, under most operating conditions, the gearbox maintains high efficiency.}
  \label{fig:experiment-actuator-efficiency}
\end{figure}

We tested the actuator across three different speeds that correspond to its typical operating range on the robot. Each torque and speed command was maintained for one second prior to measurement to minimize transient effects. As shown in Figure~\ref{fig:experiment-actuator-efficiency}, the measured efficiency trends align with previous findings~\cite{urs2022design, lee2019empirical}, demonstrating a decline in efficiency at higher torque and speed settings. 
The gearbox exhibits a mechanical efficiency of approximately 90\% across most operating conditions. However, at high torque and velocity, efficiency decreases due to heat generation. 

\subsection{Transmission Stiffness}

\begin{figure}[t]
  \centering
  \includegraphics[width=0.45\textwidth]{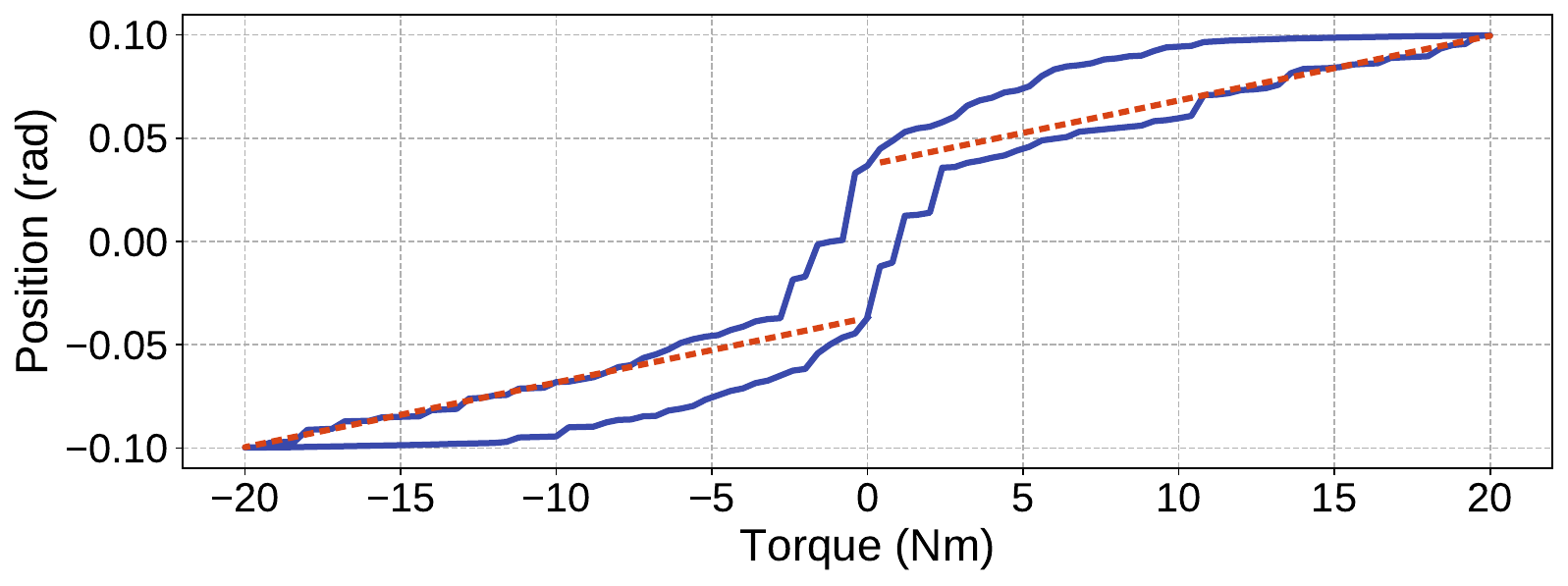}
  \caption{Transmission stiffness measurement. The output of a 6512 actuator was fixed while the motor's command torque was gradually increased. The rotor position was recorded, yielding a mean stiffness of \SI{319}{\N\m}.}
  \label{fig:experiment-actuator-stiffness} 
\end{figure}

We characterized the actuator’s transmission stiffness by rigidly fixing the output shaft relative to the actuator housing and measuring motor displacement under a range of static torques. The torque command was gradually ramped from 0~\si{Nm} to 20~\si{Nm} and back in both directions. A linear fit was then applied to the data collected from 4~\si{Nm} to 10~\si{Nm}, and the inverse of the slope yielded a stiffness of approximately 319.49~\si{Nm/rad} (Figure~\ref{fig:experiment-actuator-stiffness}). This profile shape is consistent with previous measurements of 3D-printed cycloidal reducers reported by \citet{roozing2024experimental}, who measured a substantially higher stiffness of about 1468~\si{Nm/rad} for a similar 3D-printed cycloidal reducer. The observed reduction in stiffness is likely attributable to the lower strength of PLA compared to the carbon fiber reinforced polyamide (PA-CF) material used in that study.

\subsection{Durability}

A primary concern for any 3D-printed actuator is its operational lifespan. We conducted a 60-hour durability test in which the actuator repeatedly lifted a pendulum (0.5~\si{kg}, 0.5~\si{m}) through a range of -45 degrees to +90 degrees at a frequency of 0.5~\si{Hz}. At one-hour intervals for the first 12 hours—and every 12 hours thereafter—we paused the test to measure both efficiency (using the same dynamometer stand) and backlash.

\begin{figure}[t]
    \centering
    \includegraphics[width=\linewidth]{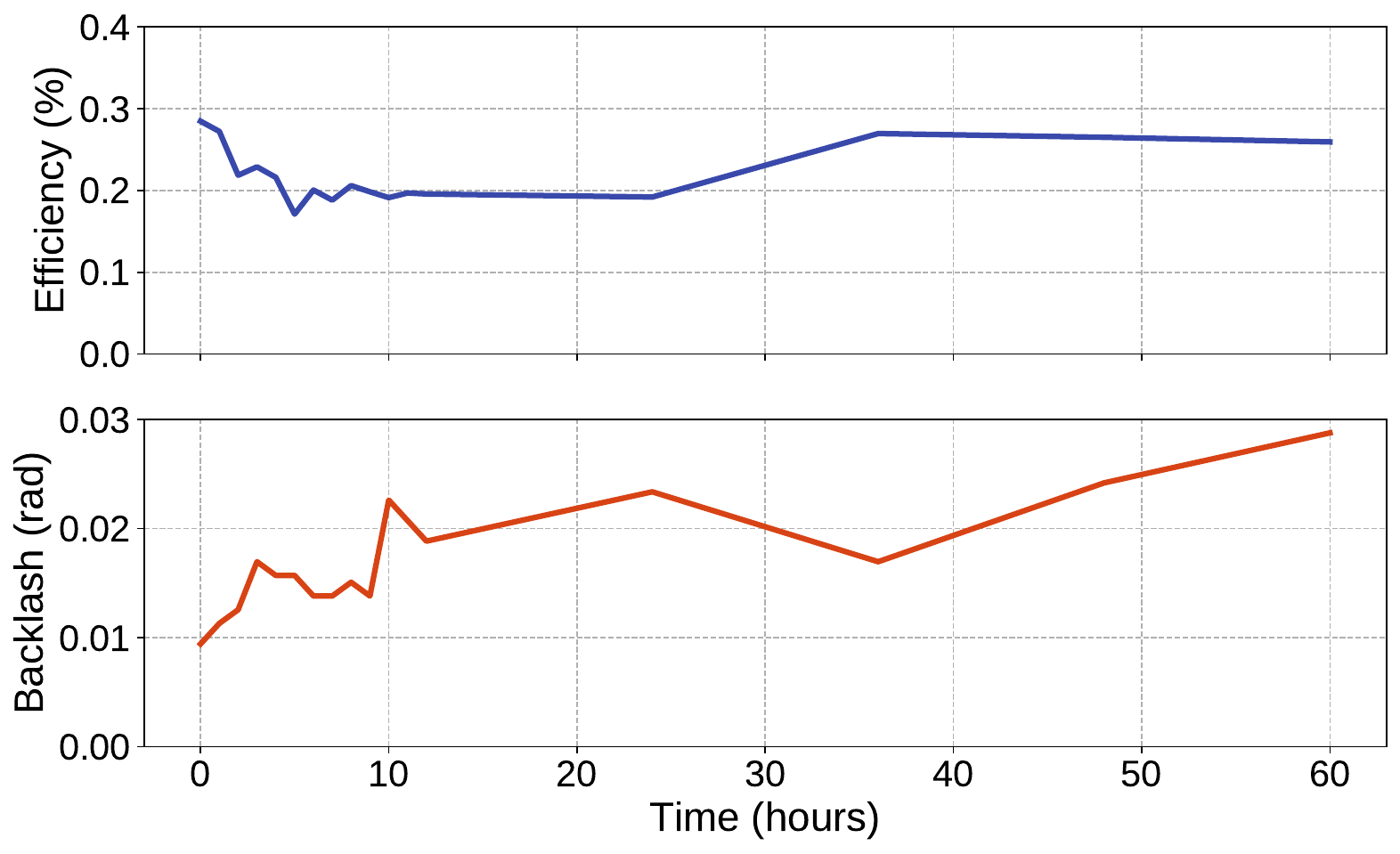}
    \caption{Average total efficiency and backlash measured during the 60-hour durability test of the 6512 actuator. The efficiency remains stable throughout the test, while backlash progressively increases due to wear.}
    \label{fig:experiment-actuator-durability} 
\end{figure}

As shown in Figure~\ref{fig:experiment-actuator-durability}, efficiency initially declined but later returned to near its original level, aligning with observations reported by \citet{urs2022design}. Although backlash increased slightly as the 3D-printed parts experienced wear, it remained within acceptable limits throughout the 60-hour test.

\subsection{Consistency Across Units}

\begin{figure}[t]
  \centering
  \includegraphics[width=0.9\linewidth]{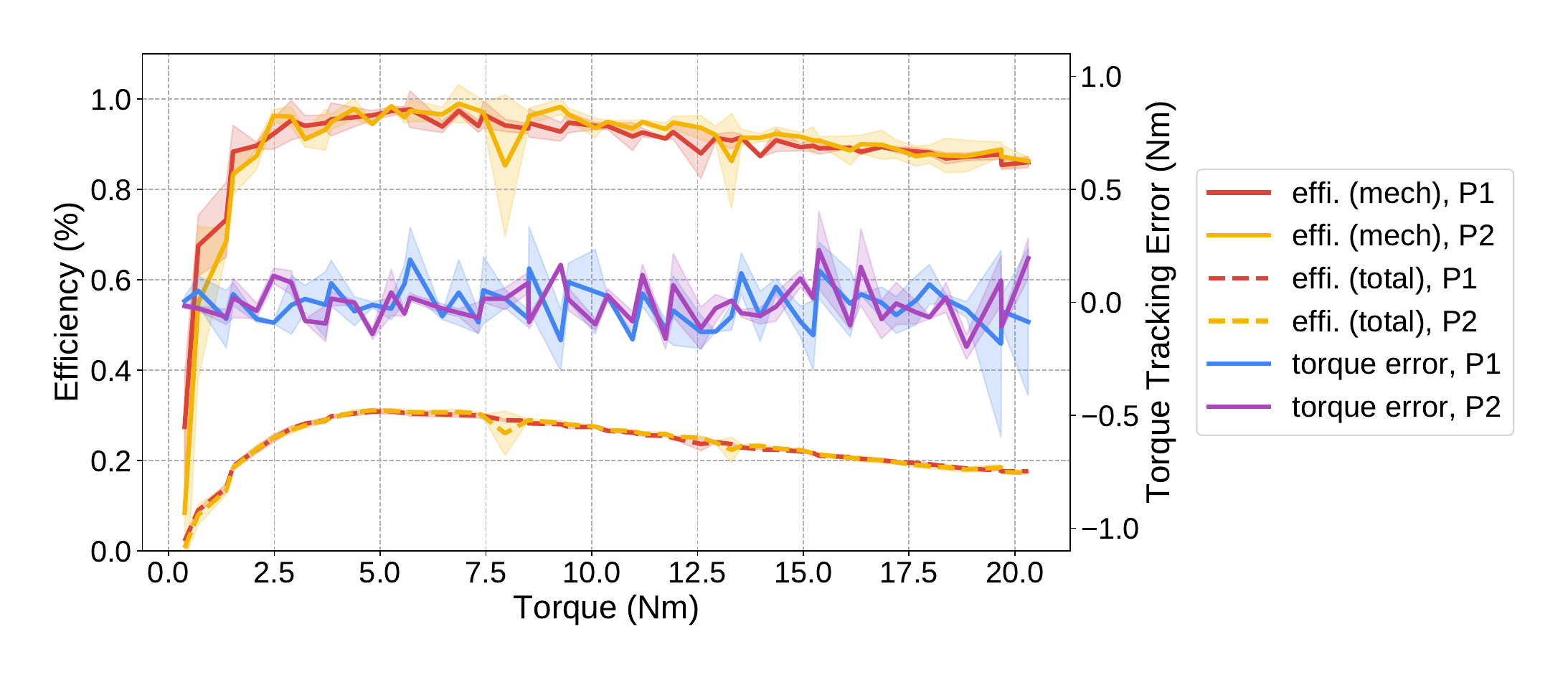}
  \caption{Average efficiency and torque tracking error at $\SI{1}{\radian\per\s}$ across six 6512 actuator samples printed on two different 3D printers (P1 and P2).}
  \label{fig:experiment-rebuttal-efficiency-torque}
\end{figure}

The performance might vary across actuator parts printed under different environment conditions and with different printers. To demonstrate the consistency of our design, we performed the efficiency evaluation on six actuators printed on two different 3D printers. The efficiency and torque tracking error, shown in Figure~\ref{fig:experiment-rebuttal-efficiency-torque}, are consistent across units. The torque error remains within $\SI{\pm0.5}{\N\m}$ throughout the operating range. This confirms low inter-sample variance and confirms that the actuator design is robust across different fabrication sources.

\subsection{Position Tracking Accuracy}

Gear‑backlash is the dominant steady‑state source of position error in our actuator stack. Therefore, we evaluate the position tracking accuracy of individual actuator modules by measuring backlash on six freshly printed 6512 actuators. The maximum observed backlash was $\SI{0.0229}{\radian}$ with a standard deviation of $\SI{0.0042}{\radian}$.

\begin{figure}[t]
  \centering
  \def\width{0.5\linewidth}
  \begin{tikzpicture}
    \matrix (m) [%
      matrix of nodes,
      inner sep=0pt,
      ampersand replacement=\&,
      anchor=south west
      ] {
      \includegraphics[width=\width]{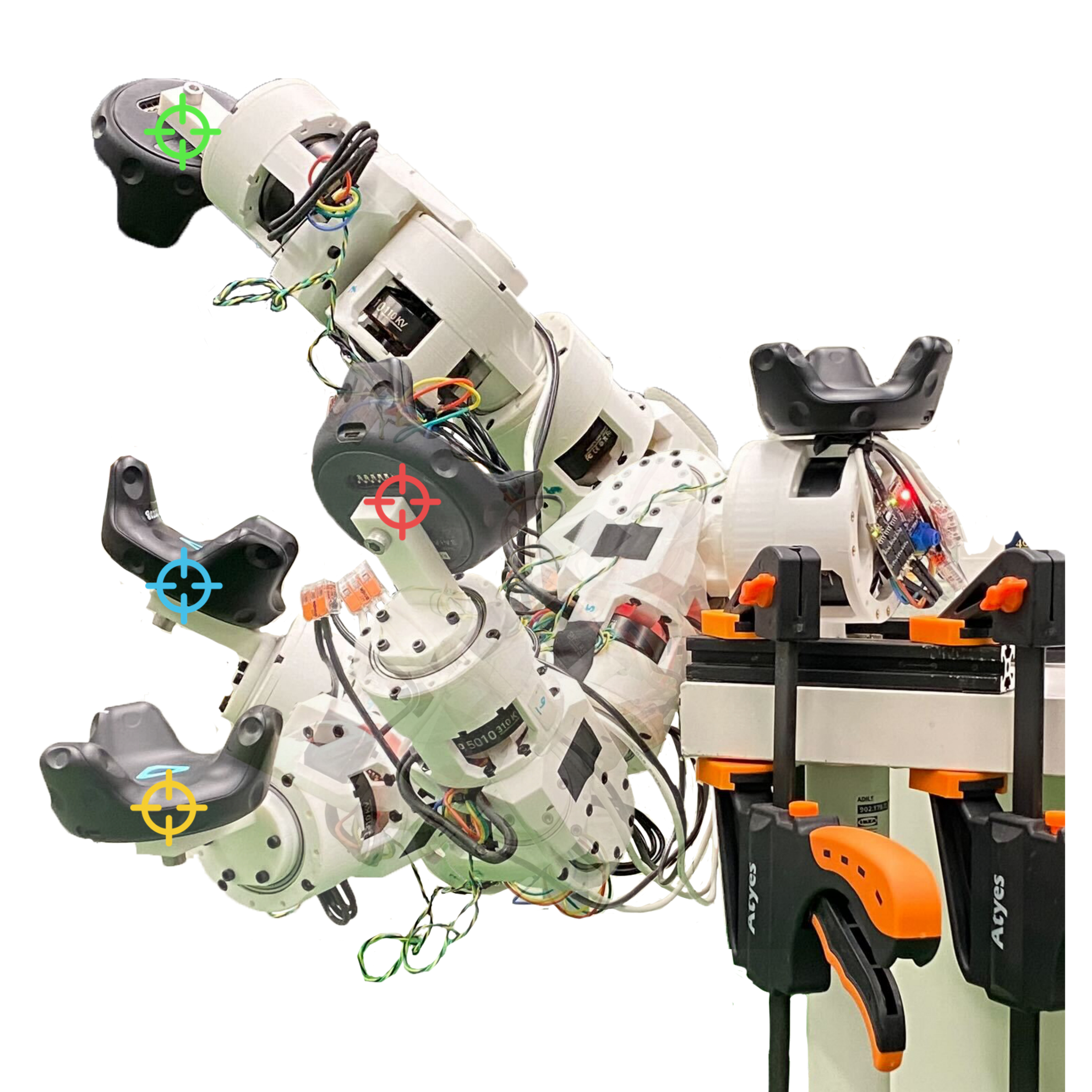} \&
      \includegraphics[width=\width]{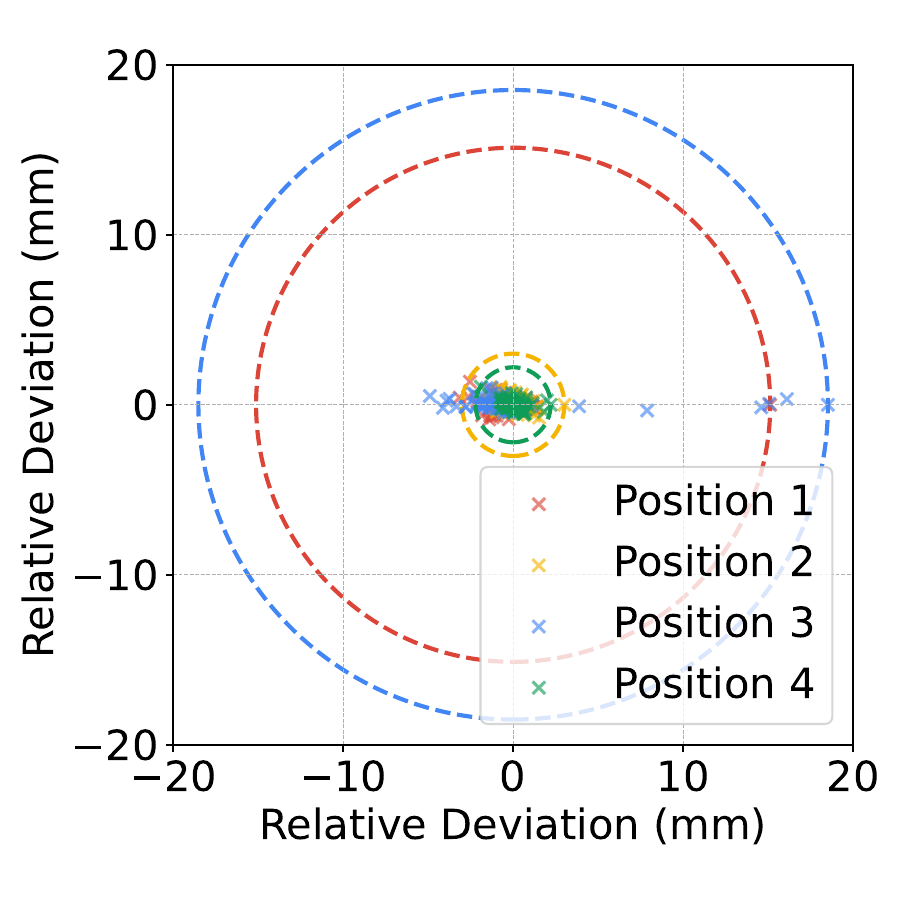} \\
      };
  \end{tikzpicture}
  \caption{(Left) The 5-DoF arm is commanded to reach four spatial position targets. (Right) End-effector position variance over 100 repetitions of the reaching task.}
  \label{fig:experiment-position-accuracy} 
\end{figure}

To evaluate the system-level performance, we assembled five actuators into a 5-DoF serial arm and commanded it to reach four spatial targets (Figure~\ref{fig:experiment-position-accuracy}). Each target was repeated $100$ times while the end‑effector pose was captured by the SteamVR-based motion‑tracking setup described in section~\ref{sec:teleoperated-manipulation}. The results show consistent positioning with standard deviation of $\SI{3.433}{\mm}$.

\begin{figure}[t]
  \centering
  \includegraphics[width=0.9\linewidth]{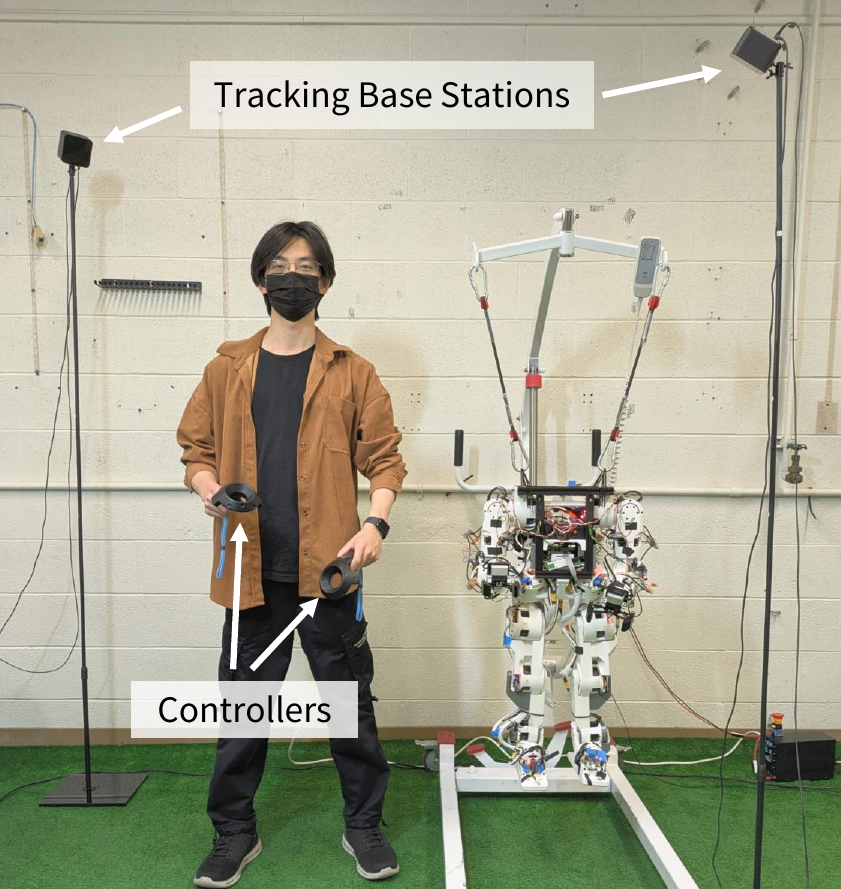}
  \caption{Teleoperation setup. Two SteamVR base stations are used to track the global position of the controller. The motion is then synchronized to the robot through our teleoperation software stack.}
  \label{fig:experiment-teleoperation-setup}
\end{figure}

\begin{figure*}[t]
  \centering
  \includegraphics[width=\textwidth, trim={0 0 0 0}, clip]{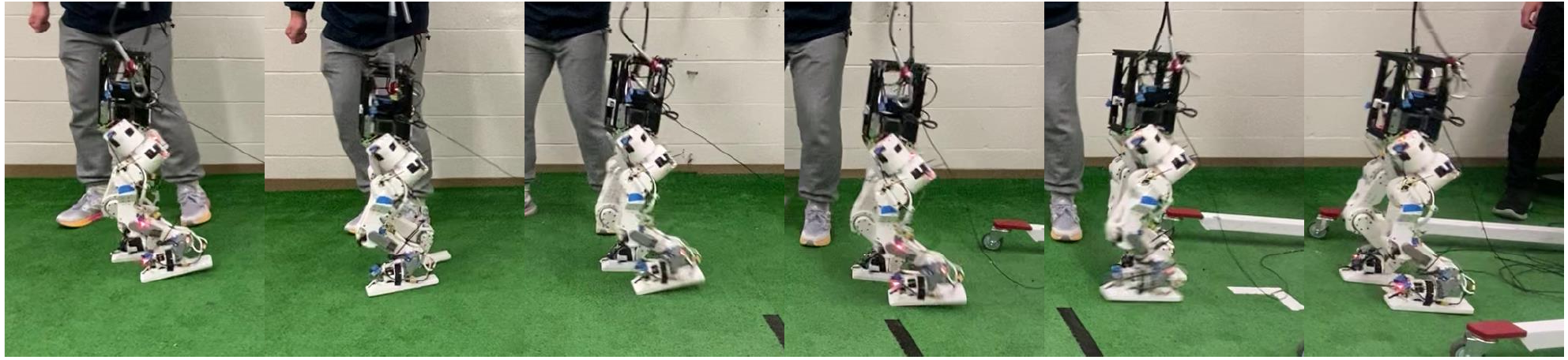}
  \caption{Snapshot of the robot in biped configuration walking forward with a trained RL policy. The robot is able to track a user commanded velocity.}
  \label{fig:experiment-locomotion}
\end{figure*}

\begin{figure*}[t]
  \centering
  \includegraphics[width=\textwidth]{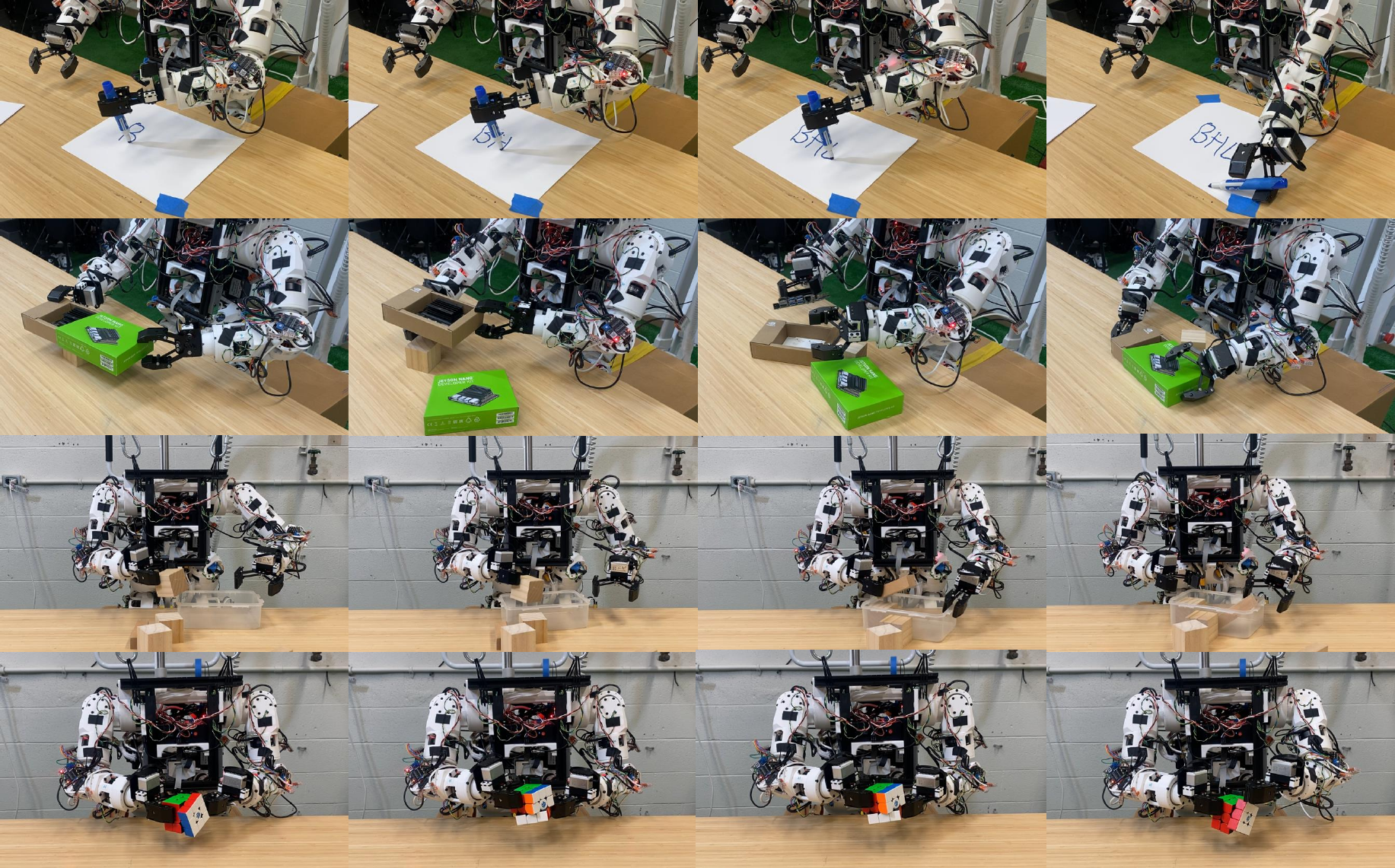}
  \caption{Snapshot of teleoperated manipulation experiments, including four tasks: writing, unpacking and packing, picking and placing blocks, and solving Rubik's Cube.}
  \label{fig:experiment-teleoperation} 
\end{figure*}

\section{Experiments}

In this section, we evaluate Berkeley-Humanoid’s ability to perform standard tasks of interest in humanoid robotics research. To this end, we conduct experiments in two key domains: locomotion and manipulation.

\subsection{Legged Locomotion}
\label{sec:locomotion}

Our design serves as a robust platform for investigating humanoid robot locomotion. Notably, we achieve a direct sim-to-real transfer of an Isaac Gym-trained policy to the physical robot without relying on additional state estimation methods. We formulate the locomotion task as a Partially Observed Markov Decision Process (POMDP) and use a standard Proximal Policy Optimization (PPO) algorithm to learn a control policy. The policy receives proprioceptive observations from the robot hardware: (a) the base angular velocity, (b) the projected gravity vector, and (c) joint positions and velocities. Additional inputs include the commanded linear velocity provided by the user and the previous time-step action. The policy outputs desired joint positions to the joint actuators. During deployment, the multi-layer perception (MLP) policy runs onboard the robot at 25~\si{Hz}. Figure~\ref{fig:experiment-locomotion} shows snapshots of the robot walking in the bipedal configuration under the trained reinforcement learning policy that follows user-specified velocity commands. Notably, the experiment utilized only 30\% of the actuator's torque limit, suggesting that the actuators are operating well within their capacity. This observation implies that the same actuator design could support a larger humanoid robot—such as the example presented in the previous section—without compromising performance.

\subsection{Teleoperated Manipulation}
\label{sec:teleoperated-manipulation}

The two 5-DOF arms and integrated grippers enable the robot to serve as an effective platform for bimanual manipulation research. To demonstrate its capabilities, we implemented a teleoperation system that leverages a motion capture setup built with SteamVR base stations and controllers. This system provides real-time tracking of the global position and orientation of the operator's hand. The SteamVR tracking system was chosen for its broad support within the OpenXR ecosystem and its cost-effectiveness. However, our open-source software stack is designed for flexibility and can be readily adapted to incorporate alternative tracking solutions, such as Apple Vision Pro, Meta Quest, OptiTrack, or Xsens, to suit a variety of research needs. Our setup is shown in Figure~\ref{fig:experiment-teleoperation-setup}.

Our teleoperation system offers two distinct operational modes: Headless Mode and VR Mode, providing flexibility for diverse user needs and environments. Both modes leverage a unified inverse kinematics (IK) pipeline based on the Pink and Pinocchio library \citep{pink2024, carpentier2019pinocchio, pinocchioweb}. Headless mode facilitates a ``third-person'' perspective, ideal for users without access to a VR display. In this configuration, the system interprets the position and orientation changes of controllers relative to the global coordinate frame. This allows the operator to intuitively control the robot by observing its actions from an external viewpoint, similar to controlling a character in a video game. In addition, the user can engage and disengage anytime, allowing for comfortable adjustment of their gesture. VR mode delivers an immersive, first-person control experience. It directly maps the position and orientation changes of controllers within the user's local coordinate frame to the robot's local coordinate frame. This ``body-ownership'' approach enables operators to perform manipulations as naturally as they would with their own hands in a real-world setting, fostering a strong sense of presence and control. Both modes utilize the same robust IK solver built upon the Pink and Pinocchio library \citep{pink2024, carpentier2019pinocchio, pinocchioweb}. This ensures consistent and reliable translation of the desired end-effector poses into corresponding robot joint configurations, regardless of the chosen operational mode. As shown in Figure~\ref{fig:experiment-teleoperation}, we are able to use this setup to (a) write with a marker (b) unpack and pack a box, (c) solve a Rubik's Cube, and (d) pick and place blocks.

\subsection{Potential Applications}

We further discuss potential extensions of Berkeley Humanoid Lite’s components, particularly in engineering education and animatronics. Given its open-source and accessible modular design, we anticipate a broader range of applications emerging with contributions from the robotics community.

\subsubsection{Education}

The 3D-printed actuator is an effective teaching tool for engineering education. In mechatronics classrooms, its accessibility enables students to gain hands-on experience with motor actuator components, including gear reduction, the motor, and its driver board. Notably, the 6512 actuator is already utilized in the UC Berkeley MECENG 102B class to demonstrate the working principle of a cycloidal drive. Its open-source and easily fabricated design allows students to experiment with different gear types beyond cycloidal gears or use modular actuators to build their own robots.
In control courses, students can develop their own FOC algorithms based on motor models or use the actuator to create an inverted pendulum testbed for controller evaluation. In robotics courses, they can conduct hands-on experiments with existing robots to implement class-taught algorithms, such as motion planning, reinforcement learning, or imitation learning. 
The system’s modular, accessible, and easily repairable design also benefits instructors, allowing for quick replacement of broken parts. 
Beyond higher education, this platform can also inspire and engage middle and high school students in robotics.

\subsubsection{Animatronics}

\begin{figure}[t]
  \centering
  \includegraphics[width=0.9\linewidth]{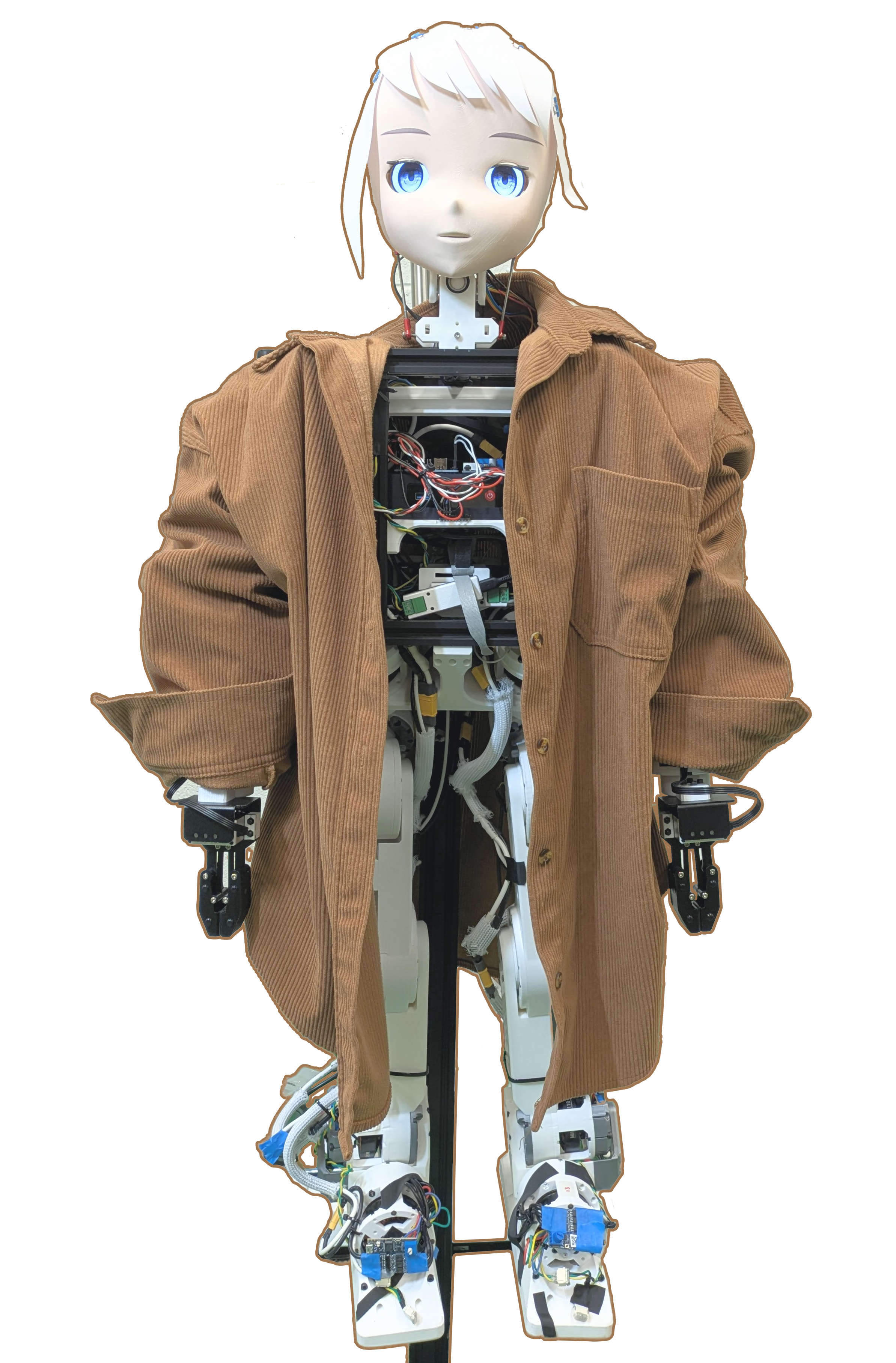}
  \caption{An anime-style head is integrated into the robot, enabling it to embody expressive and visually appealing anime characters while enhancing its overall aesthetic and potential for engaging human-robot interaction.}
  \label{fig:animatronics}
\end{figure}

Due to its modularity and lightweight design, the Berkeley Humanoid Lite can also serve as an animatronics platform for entertainment, as demonstrated in Figure~\ref{fig:animatronics}. 
By refining its form to enhance expressiveness—incorporating features such as cuteness and vividness~\cite{al2021anime}—the robot can achieve a more engaging and visually appealing presence. Moreover, employing body motion design techniques inspired by animation can generate expressive movements, which the physical robot can execute using the control methods detailed in~\cite{li2020animated,grandia2025design}. With its low cost, ease of fabrication and straightforward repairability, Berkeley Humanoid Lite presents a promising platform for entertainment applications, particularly in settings aimed at young audiences.

\section{Limitations}

One limitation of this work is the insufficient study of thermal effects on the 3D-printed structure during prolonged operation, which could impact both mechanical strength and overall system reliability. Future iterations will include more rigorous thermal testing and structural refinements to address these concerns. Beyond the robot morphologies presented in this work, there are more potential new morphologies that can be explored and made openly available.

\section{Conclusion}

In this work, we introduced an affordable, mid-scale humanoid robot platform that significantly lowers the barriers to entry for researchers, educators, and hobbyists alike. By leveraging 3D-printed cycloidal actuators, off-the-shelf components, and a fully open-source hardware and software stack, we demonstrated that effective humanoid systems need not rely on proprietary, high-cost solutions. We evaluated the reliability and versatility of the 3D-printed actuators and demonstrated the ability of the robot platform to perform locomotion and teleoperation tasks. By sharing all design files, control algorithms, and manufacturing details, we hope to democratize access to humanoid robot hardware, fostering broader participation in humanoid-robotics research and accelerating innovations in this critical domain.

\section{Acknowledgment}

We would like to thank Lydia Liu, Widyadewi Soedarmadji, and Daniel Wong for the early-stage project explorations. We would also like to thank Alex Hao and Ted Zhang for providing help on supporting the experiments. We are grateful to Chengyi Lux Zhang for the generous assistance. Finally, we appreciate the helpful discussions from all members of Hybrid Robotics Group and SLICE lab. This work is supported in part by NSF 2303735 for POSE, in part by NSF 2238346 for CAREER, in part by the Robotics and AI Institute. K. Sreenath has financial interest in the Robotics and AI Institute. He and the company may benefit from the commercialization of the results of this research.

\bibliographystyle{plainnat}
\bibliography{references}

\end{document}